\documentclass{svproc}
\usepackage{url}
\usepackage{amsmath}
\usepackage{graphicx}

\begin{document}
\mainmatter              
\title{A Machine-Synesthetic Approach To DDoS Network Attack Detection
\thanks{This report contains results of the research project supported by Russian Foundation for Basic Research, grants no. 18-07-01109, 16-47-330055}}
\titlerunning{}  
%
\author{Yuri Monakhov \and Oleg Nikitin\and
Anna Kuznetsova \and Alexey Kharlamov \and Alexandr Amochkin}
\authorrunning{Yuri Monakhov et al.} 
%
\tocauthor{Yuri Monakhov, Oleg Nikitin, Anna Kuznetsova, Alexey Kharlamov, Alexandr Amochkin}
\institute{Vladimir State University, 600000 Vladimir, Russia
\email{unklefck@gmail.com}}

\maketitle              

\begin{abstract}
In the authors’ opinion, anomaly detection systems, or ADS, seem to be the most perspective direction in the subject of attack detection, because these systems can detect, among others, the unknown (zero-day) attacks. To detect anomalies, the authors propose to use machine synesthesia. In this case, machine synesthesia is understood as an interface that allows using image classification algorithms in the problem of detecting network anomalies, making it possible to use non-specialized image detection methods that have recently been widely and actively developed. The proposed approach is that the network traffic data is ”projected” into the image.It can be seen from the experimental results that the proposed method for detecting anomalies shows high results in the detection of attacks. On a large sample, the value of the complex efficiency indicator reaches 97\%.
\end{abstract}
\section{Introduction}

One of the methods of ensuring network availability is employing the
network anomaly detection mechanisms.Before defining an anomaly, it is
necessary to figure out what is considered a normal state. We consider the state of system "normal" (or "functionally viable") when it performs all the functions assigned to it. Therefore, an anomaly is a state where the behavior of the system does not correspond to
the clearly established characteristics of normal behavior \cite{mohiuddin}.
Implementing the prompt detection mechanisms for such anomalies will sufficiently increase the chances of an effective response to network availability violation incidents.

Known network anomalies are so diverse they cannot be categorized using a single classification. There is a clearly laid distinction, however, between active and passive, external and internal, intentional and unintentional anomalies, etc. Since these distinctions do not reflect all the characteristics of the phenomenon under study, the author \cite{afontsev} proposed a classification of anomalies based upon the impact object, i.e. an information system consisting of hardware, software and a network infrastructure.

According to the chosen approach, network anomalies can be divided into two main groups: node malfunctions and security breaches. Node malfunctions include hardware faults, design and configuration errors, software errors, and hardware performance issues. Network security breaches include the following anomalies: network scanning, denial of service, malware activity, distribution of network worms, exploitation of vulnerabilities, traffic analyzers (sniffers), and network modifiers (packet injections, header spoofing etc).

The largest financial damage to telecom operators is caused by denial of
service incidents. DoS attacks, in turn, can be divided into two types: inadvertently caused "attacks" (design errors and network settings, a small amount of dedicated computing resources, a sharp increase in the number of calls to a network resource) and attacks due to deliberate actions, e.g. UDP flood, TCP SYN flood, {\it Smurf} ICMP broadcast flood and ICMP flood. Deliberate attacks pose the greatest threat, as it is more difficult to mitigate them effectively and potentially they can lead to large losses.

Analysis of research results published in \cite{berestov,galtsev,kornienko,kussul,mirkes,tsvirko}, as well as reports of major information security systems developers, showed that there is no single effective algorithm for denial-of-service attack detection and mitigation. Usually, vendows offer an expensive solution implementing a hybrid algorithm based on signature search methods and blacklisting attacker node IP addresses as a form of mitigation. An example is the ATLAS system from Arbor, Ltd. Thus, the problem of developing tools for distributed DoS attack detection with a high degree of efficiency remains relevant.

The rest of this paper is organized as follows: in Section 2, a review of the existing approaches for detecting anomalies is provided; Section 3 discusses the proposed approach, specifically, a strategy of representing traffic metadata into an image and an algorithm for classifying the obtained image are presented; in Section 4, experimental results are provided; Section 5 concludes the work and gives an outlook for further studies.

\section{Existing approaches}

In the authors' opinion, anomaly detection systems, or ADS, seem to be the most perspective direction in the subject of attack detection, because these systems can detect, among others, the unknown ({\it zero-day}) attacks.
Almost all the models for detecting anomalies described in the literature can be divided into:
\begin{itemize}
	\item [a)] based on a behavioral pattern storage \cite{somayaji,ilgun}. The program implementation of this approach needs to be compiled into the operating system kernel, which is difficult to the point of practical impossibility (e.g. in trusted computing systems). In addition, the constant presence of a monitoring component leads to an overall slowdown of the entire system by approximately 4 - 50 percent;
	\item [b)] frequency-based \cite{eskin:lee,ye:xu}. Common drawbacks of frequency methods are their poor adaptability, since the reference values of frequencies are determined once by training sets or according to expert data. Moreover, these methods are usually "stateless", i.e. the order of feature appearance is not taken into account;
	\item [c)] based on some type of a neural network classifier \cite{michael,garvey,theus,tan,ilgun:kemmerer}. The disadvantage of many neural networks is their poor fitness to process non-ordered datasets. Introducing an artificial order on a set of element values will only distort the picture, since the neural network will recalculate weights according to the proximity of numerical values;
	\item [d)] based on a finite automata (state machine) synthesis \cite{kussul,somayaji,eskin,ghosh,ye}. The main disadvantage of this approach is the complex process of building a state machine by parsing the attack scenario. In addition, there are restrictions on the types of attack algorithms that can be described by regular grammars;
	\item [e)] other, special: based on Bayesian networks \cite{axelsson}, genetic algorithms \cite{chikalov}, etc. Most of the works offer only the basic idea, the algorithm, often unsuitable for practical use.
\end{itemize}

\section{Proposed approach}

To detect anomalies, the authors propose to use machine synesthesia. In this case, machine synesthesia is understood as an interface that allows using image classification algorithms in the problem of detecting network anomalies, making it possible to use non-specialized image detection methods that have recently been widely and actively developed \cite{chen}. The proposed approach is that the network traffic data is "projected" into the image. Accumulating inage chanfes gives us a video stream, analyzing which, we can make a conclusion about the anomalous state of the observed data network.

The basis of any anomaly detection system is a module that analyzes network packets and decides on their potential maliciousness. In fact, ADS is trying to classify network traffic into two subsets: “normal” traffic and network attacks (it doesn't even matter which detection technology is used — signature-based or statistical). Consequently, the very concept of ADS is in very good agreement with the goals of image classification algorithms - matching the original image to a class of images from a set according to some features. Moreover, image classification as a mathematical tool for analyzing network traffic data and detecting network attacks has several advantages compared to the anomaly detection methods discussed earlier. These advantages are represented below.
\begin{itemize}
	\item The mathematical apparatus for the classification of images is well developed and tested in practice in many other areas of science and technology.
	\item A large number of image classification algorithms and wide possibilities for their improvement make this mathematical apparatus very flexible and provide an extensive potential for increasing the efficiency of network intrusion detection.
	\item Most image classification algorithms, showing high practical efficiency, are relatively easy to understand and implement in software.
	\item Image classification is very effective even with very large amounts of input data. This fact makes us consider these methods as especially suitable for analyzing large network traffic dumps.
	\item Classification of images can be applied even in the absence of a priori information about the importance of particular network packet features in the context of detecting certain types of network attacks.
	\item Interpretation of the results is fairly simple and intuitive.
\end{itemize}
\subsection{Image representation of multidimensional TCP/IP traffic data}

The authors propose to solve the problem of representing network traffic metadata in the way which will allow using the pattern recognition algorithm to detect anomalies in the video stream.

Consider the network terminal device collecting traffic in the virtual channel. Each collected packet has a set of metadata, presented as a 
vector $p$:
$$ p(id, date, x_1, x_2, \dots, x_n), p \in P, $$

where $n$ is a vector dimension, $P$ is a set of all vectors, $id$ is a session identifier, $date$ is a timestamp of logging by the terminal, $x_1, \dots, x_n$ - direction, adresses and ports of sender and receiver, packet size, protocol type, timestamp (as in TCP segment header), different flags and service fields.

To project traffic into an image, the “orthogonal projection” method is used \cite{gantmacher}: each vector p is represented by a point in multidimensional space, where n is the dimension of space, then all points (packets) belonging to one session are projected into two-dimensional space:
$$
X' = \frac{
	\begin{vmatrix} 
		(\bar{a}\bar{a}) & (\bar{a}\bar{b}) & a \\
		(\bar{b}\bar{a}) & (\bar{b}\bar{b}) & b \\
		\bar{X}\times\bar{a} & \bar{X}\times\bar{b} & 0
	\end{vmatrix}
}
{
	\begin{vmatrix} 
		(\bar{a}\bar{a}) & (\bar{a}\bar{b})\\
		(\bar{b}\bar{a}) & (\bar{b}\bar{b})
	\end{vmatrix}
}
$$

where $a$, $b$ are empirically achieved basis vectors for the projection into the two-dimensional space, $X$ is a source vector, constructed from $p$ by removing $id$ and $date$ elements, $X'$ is a projection result, $\times$ is a cross product, $()$ is a scalar product.

The next stage of the network session imaging is the connection of all its points, forming a convex figure. The last step is to fill the resulting shape with color. Then everything is repeated for the next network session. The resulting image is obtained when the imaging process has been performed for all network sessions intercepted by the terminal. Accumulating changes or differentiating this image gives us a video stream. Fig. 1 shows examples of images that reflect the legitimate ("normal state") network behavior.
\begin{figure*}[h]
\centerline{\includegraphics[width=0.8\textwidth]{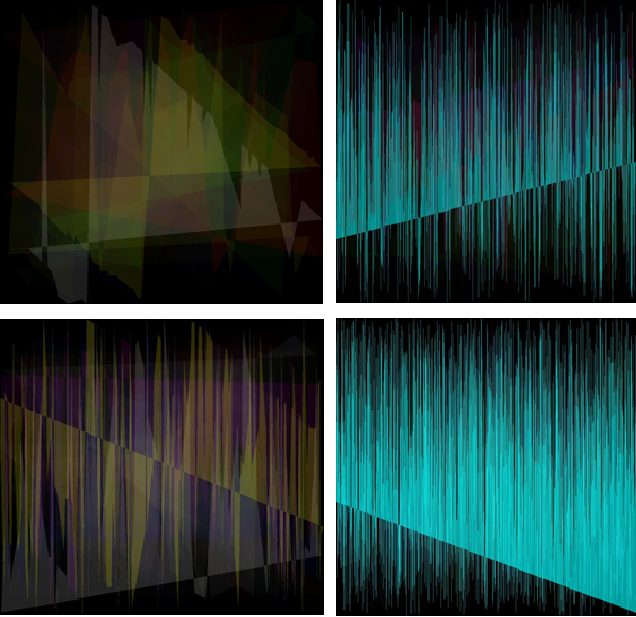}}
\caption{}
\label{fig1}
\end{figure*}

\subsection{Image classification in the problem of anomaly detection}

The next step is to solve the problem of classifying the obtained image.
In general, the solution to the task of detecting classes (objects) in an image is to use machine learning algorithms for building class models, and then output algorithms to search for classes (objects) in an image.

Building a model has two stages:
\begin{itemize}
	\item [a)] Extraction of characteristic features for a class: construction of characteristic feature vectors for class elements.
	\item [b)] Training on the obtained features of the model for subsequent recognition tasks.
\end{itemize}

The description of the class object is carried out using feature vectors. Vectors are built from:
\begin{itemize}
	\item [a)] color information (oriented gradient histogram);
	\item [b)] contextual information;
	\item [c)] data on the geometric interposition of object parts.
\end{itemize}

The classification (prediction) algorithm can be divided into two stages:
\begin{itemize}
	\item [a)] Extracting features from an image. At this stage, two tasks are performed:
	\begin{itemize}
		\item Since the image can contain objects of many classes, we need to find all the representatives. To do this,one might use a sliding window, which “runs through” the image from the upper left to the lower right corner.
		\item The image is scaled, since the scale of the objects in an image may vary.
	\end{itemize}
	\item [b)] Associating an image with a specific class. A formal class description, i.e. a set of features that are highlighted by their test images, is used as an input data. Based on this information, the classifier decides whether the image belongs to the class and assesses the degree of reliability for the conclusion.
\end{itemize}

\subsubsection{Classification methods.}
Classification methods range from mostly heuristic approaches to formal procedures based on the methods of mathematical statistics. There is no generally accepted classification, but a number of approaches to image classification can be distinguished:
\begin{itemize}
	\item methods of part-based object modeling;
	\item "bag-of-words" methods;
	\item spatial pyramid matching methods.
\end{itemize}
For implementation presented in this article the authors chose the bag-of-words algorithm, considering the following reasons:
\begin{itemize}
	\item The algorithms of the parts-based modeling and spatial pyramid matching are sensitive to the position of the descriptors in space and their mutual arrangement. These classes of methods are effective in the tasks of detecting objects in an image; however, due to the characteristic features of the input data, they are poorly applicable to the problem of image classification.
	\item The bag-of-words algorithm is widely tested in other areas of knowledge, it shows good results and is simple enough to implement.
\end{itemize}
To analyze the video stream projected from the traffic, we used a naive Bayes classifier \cite{murty}. It is often used to classify texts with the bag-of-words model. In this case, the approach is similar to the analysis of texts, only descriptors are used instead of words. The work of this classifier can be divided into two parts: the training phase and the prediction phase.

\subsubsection{Training phase.}
Each frame (image) is fed to the input of the descriptor search algorithm, in this case the scale-invariant feature transform (SIFT)\cite{lowe}. After that, the task of correlating singular points between frames is performed. A singular point on the image of an object is a point that will most likely appear on other images of this object.

To solve the problem of comparing the singular points of an object in different images, a descriptor is used. Descriptor is a data structure, identifier of a singular point, distinguishing it from the rest. It may or may not be invariant w.r.t. image transformations of the object. In this case, the descriptor is invariant w.r.t. perspective transformations, i.e. scaling. The descriptor allows to compare a singular point of the object in one image with the same singular point on another image of this object.

Next, the set of descriptors obtained from all images is ordered into groups “by similarity” using the k-means clustering method \cite{lowe, kmeans}. This is done in order to train the classifier, which will issue a conclusion about whether the image represents anomalous behavior.

Below is a step-by-step algorithm for training the image descriptor classifier:

\begin{itemize}
	\item [\textbf{Step 1.}]  Extraction of all descriptors from sets with attack and without attack.
	\item [\textbf{Step 2.}] K-means clustering of all descriptors into n clusters.
	\item [\textbf{Step 3.}] Calculation of the matrix $A(m, k)$, where $m$ is the number of images and $k$ is the number of clusters. The element $(i; j)$ will store the value of how frequently the descriptors from the $j$-th cluster appears on the $i$-th image. Such a matrix will be called the appearance frequency matrix.
	\item [\textbf{Step 4.}] Calculation of descriptor weights using $tfidf$ formula \footnote
	{Wu H., Luk R., Wong K., and Kwok K.: Interpreting TF-IDF term weights as making relevance decisions. ACM Transactions on Information Systems, vol. 26, no. 3, (2008)}
		$$ tfidf(t, d, D) = tf(t, d)*idf(t, D)$$
Here $tf$ ("term frequency") is the frequency of occurrence of the descriptor in this image and is defined as
		$$ tf(t, d) = \frac {n_t} {\sum\limits_k n_k},$$
where $t$ is a descriptor, $k$ is the number of descriptors in an image, $n_t$ is an amount of descriptor $t$ in an image.
Also, $idf$ ("inverse document frequency") is the reverse frequency of the image with the given descriptor in the sample and is defined as
		$$ idf(t, D) = log\frac{D}{\{d_i \in D, t \in d_i\}}, $$
where $D$ is the number of images with the given descriptor in the sample,$\{d_i \in D, t\in d_i\}$ is the number of images in $D$ where $t$ is found under the conditions of $n_t \neq 0$.
	\item [\textbf{Step 5.}] Substituting corresponding weights instead of descriptors into the matrix $A$.
	\item [\textbf{Step 6.}] Classification. We use the “boosting” (adaboost) of naive Bayes classifiers.
	\item [\textbf{Step 7.}] Saving the trained model to a file.
	\item [\textbf{Step 8.}] THis concludes the training phase.
\end{itemize}

\subsubsection{Prediction phase.}
The differences between the training phase and the prediction phase are small: descriptors are extracted from the image and related to the groups at hand. Based on this relationship, a vector is constructed. Each element of this vector is the frequency of occurrence of descriptors from this group in the image. Analyzing this vector, the classifier can make a prediction about an attack with a certain probability.

General algorithm for prediction based on a pair of classifiers is presented below.
\begin{itemize}
	\item [\textbf{Step 1.}] Extraction of all descriptors from the image;
	\item [\textbf{Step 2.}] Clustering the resulting set of descriptors;
	\item [\textbf{Step 3.}] Calculation of the vector $[1, k]$;
	\item [\textbf{Step 4.}] Calculation of the weight for each descriptor by the $tfidf$ formula presented above;
	\item [\textbf{Step 5.}] Replacing the frequency of occurrence in vectors with their weight;
	\item [\textbf{Step 6.}] Classification of the resulting vector by a previously trained classifier;
	\item [\textbf{Step 7.}] Conclusion about the presense of an anomaly in the observed network based on the prediction of the classifier.
\end{itemize}
\section{Detection efficiency evaluation}

The task of evaluating the efficiency of the proposed method was solved experimentally. The experiment used a number of parameters set empirically. 1000 clusters were used for clustering. The generated images were 1000 by 1000 pixels.
\subsection{Experimental dataset}

A setup was assembled for the experiments. It consists of three devices connected by a communication channel. The block diagram of the setup is shown in Figure 2.

\begin{figure}[htbp]
\centerline{\includegraphics[width =0.6\linewidth]{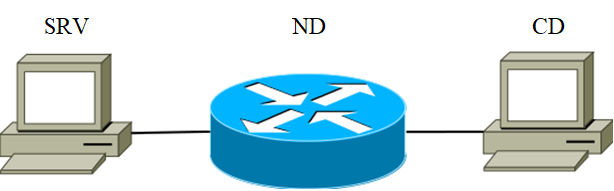}}
\caption{}
\label{fig2}
\end{figure}

The SRV device plays the role of a server under attack (hereinafter referred to as the target server). As the target server, the devices listed in Table 1 with the code SRV were used sequentially. The second is a network device designed to transfer network packets. Characteristics of the device are shown in Table 1 under the code ND-1.

\begin{table}
\caption{Network device characteristics}
\begin{center}
\begin{tabular}{ | l | p{2cm} | p{1cm} | p{2cm} | p{1cm} | p{1cm} | p{2cm} |}
\hline
{Code} & Description & RAM, MB & Network interface, Mbps & Disk drive, GB & Disk type & Processor \\
\hline
SRV-1 & Acer Atom Nettop & 2048 & 100 & 60 & SSD & 2x2GHz Intel Atom D525 \\
SRV-2 & virtual host & 6144 & 100 & 70 & SSD & 8x2GHz Intel Xeon E7-4850  \\
SRV-3 & virtual host (KVM) & 512 & 1000 & 10 & HDD & 2GHz QEMU Virtual CPU \\
ND-1 & WR842ND router & 32 & 100 & .008 & Flash & 535MHz MIPS 74Ks \\
\hline
\end{tabular}
\end{center}
\end{table}

On target servers, network packets were captured to a PCAP file for later use in detection algorithms. For this task, the tcpdump utility was used. The datasets are described in table 2.

\begin{table}
\caption{Sets of captured network packets}
\begin{center}
\begin{tabular}{ | l | p{3cm} | p{1cm} | p{1cm} | p{1cm} | p{2cm} | p{2cm} |}
\hline
{Code} & Filename & Server & DDoS & Time of record, min & No. of packets & Dump size \\
\hline
	D1 & calm\_network & SRV-1 & No & 71 & 2950108 & 2.8Gb \\
	D2 & empty\_net\_247 & SRV-2 & No & 71 & 87306 & 15Mb \\
	D3 & empty\_net & SRV-3 & No & 17 & 163950 & 11Mb \\
	D4 & pretty\_loaded & SRV-3 & Yes & 13 & 53244 & 54Mb \\
	D5 & loaded & SRV-1 & Yes & 12 & 2949244 & 433Mb \\
	D6 & loaded\_2 & SRV-2 & Yes & 5 & 589706 & 403Mb \\

\hline
\end{tabular}
\end{center}
\end{table}

The following software was used on target servers: Linux distribution, nginx 1.10.3 web server, postgresql 9.6 DBMS. To emulate system load a special web application was written. The application requests a database with a large amount of data. The request is designed to minimize the use of various caching. Through the experiments the requests to this web application were generated.

The attack was generated from the third client device (Table 1) using the Apache Benchmark utility. The structure of the background traffic  during the attack and during the rest of the time is presented in Table 3.

\begin{table}
\caption{Background traffic features}
\begin{center}
\begin{tabular}{ | l | l | l |}
\hline
	Code & Protocols & Traffic datasets\\
\hline
	BT-1 & bittorrent & D1, D5 \\
	BT-2 & ssh & All datasets \\
	BT-3 & http & D1, D3, D4, D5 \\
	BT-4 & https & D2, D6 \\
\hline
\end{tabular}
\end{center}
\end{table}
As an attack we implement a version of the HTTP GET-flood distributed DoS. Such an attack is essentially a generation of constant stream of GET requests, in this case from the CD-1 device. To generate it, we used the {\it ab} utility from the {\it apache-utils} package. As a result, files containing information about the state of the network were obtained. The main features of these files are presented in table 2. The main parameters of the attack scenario are listed in table 4.

\begin{table}
\caption{DDoS attack features}
\begin{center}
\begin{tabular}{ | l | l | l | l | l |}
\hline
	Code & Dataset code & Requests processed & Speed, pps & Avg. processing time, ms \\
\hline
	A-1 & D4 & 900 & 15.90 & 29201 \\
	A-2 & D5 & 8300 & 24.45 & 18120 \\
	A-3 & D6 & 9950 & 31.20 & 16023 \\
\hline
\end{tabular}
\end{center}
\end{table}
 
From the resulting network traffic dump, sets of the generated images TD\#1 and TD\#2, which were used for the training phase, were obtained. The sample TD\#3 was used for the prediction phase. A summary of the test datasets is presented in Table 5.

\begin{table}
\caption{Test image datasets}
\begin{center}
\begin{tabular}{ | l | l | l | l |}
\hline
	Image type & Test data TD\#1 & Test data TD\#2 & Test data TD\#3 \\
\hline
	Legitimate & 1500 images & 3000 images & 1000 images \\
	With DDoS & 500 images & 1500 images & 1000 images \\
	\hline
	{\bf Total} & 2000 images & 4500 images & 2000 images \\
\hline
\end{tabular}
\end{center}
\end{table}

\subsection{Efficiency criteria}

The main parameters evaluated through the course of this research were:
\begin{itemize}
	\item [a)] $DR$ (Detection Rate) - the number of detected attacks in relation to the total number of attacks. The higher this parameter, the higher the efficiency and quality of ADS.
	\item [b)] $FPR$ (False Positive Rate) - the number of "normal" objects, mistakenly classified as an attack, in relation to the total number of "normal" objects. The lower this parameter, the higher the efficiency and quality of the anomaly detection system.
	\item [c)] $CR$ (Complex rate) is a complex indicator that takes into account the combination of DR and FPR parameters. Since, as part of the study, the $DR$ and $FPR$ parameters were taken to be of equal importance, the complex indicator was calculated as follows: $CR = \frac{DR+FPR}{2}$.
\end{itemize}

The classifier was fed 1000 images marked as "anomalous". Based on the recognition performance, $DR$ was calculated depending on the size of the training sample. The following values were obtained: for TD\#1 $DR = 9.5\%$ and for TD\#2 $DR = 98.4\%$. Next, the second half of the images (the "normal" ones) were classified. Based on the result, $FPR$ was calculated (for TD\#1 $FPR = 3.2\%$ and for TD\#2 $FPR = 4.3\%$). Thus, the following comprehensive efficiency indicators were obtained: for TD\#1 $CR = 53.15\%$ and for TD\#2 $CR = 97,05\%$.

\section{Conclusions and future research}

It can be seen from the experimental results that the proposed method for detecting anomalies shows high results in the detection of attacks. E.g., on a large sample, the value of the complex efficiency indicator reaches 97\%. However, this method has some limitations in its application:

\begin{enumerate}
	\item The  values of $DR$ and $FPR$ show the sensitivity of the algorithm to the size of the training set, which is a conceptual problem for machine learning algorithms. Increasing the sample results in improved detection rates. However, it is not always possible to implement a sufficiently large training sample for a specific network.
	\item The developed algorithm is deterministic, the same image is classified each time with the same result.
	\item The efficiency indicators of the approach are good enough for proof of concept, but the number of false positives is also large, which can lead to the difficulties of practical implementation.
\end{enumerate}

To overcome the limitation described above (item 3), it is supposed to change the naive Bayesian classifier to a convolutional neural network, which, according to the authors, should lead to an increase in the accuracy of the anomaly detection algorithm.

%


\begin{thebibliography}{6}
%
\bibitem {mohiuddin}
Mohiuddin, A., Abdun, N.M., Jiankun, H.: A survey of network anomaly detection techniques. Journal of Network and Computer Applications, vol. 60, pp. 19?31 (2016) 
\bibitem {afontsev}
Afontsev, E.: Network anomalies, 2006  \url{https://nag.ru/articles/reviews/15588/setevyie-anomalii.html}
\bibitem {berestov}
Berestov, A.A.: Architecture of intelligent agents based on a production system to protect against virus attacks on the Internet. In: XV All-Russian Scientific Conference Problems of Information Security in the Higher School System, pp. 180?276 (2008)
\bibitem {galtsev}
Galtsev, A.V.: System analysis of traffic to identify anomalous network conditions: The thesis for the Candidate Degree of Technical Sciences, Samara (2013)
\bibitem {kornienko}
Kornienko, A.A., Slyusarenko, I.M.: Intrusion Detection Systems and Methods: Current State and Direction of Improvement, 2008 \url{http://citforum.ru/security/ internet/ids_overview/}
\bibitem {kussul}
Kussul, N., Sokolov, A.: Adaptive anomaly detection in the computer systems users behavior using Markov chains of variable order. Part 2: Methods of detecting anomalies and the results of experiments. In: Informatics and Control Problems, no. 4, pp. 83?88 (2003) 
\bibitem {mirkes}
Mirkes, E.M.: Neurocomputer: draft standard, pp. 150?176. Science, Novosibirsk (1999)
\bibitem {tsvirko}
Tsvirko, D.A. Prediction of a network attack route using production model methods, 2012 \url{http://academy.kaspersky.ru/downloads/academycup_participants/cvirko__d.ppt}
\bibitem {somayaji}
Somayaji, A.: Automated response using system-call delays. In: USENIX Security Symposium 2000, pp. 185?197 (2000) 
\bibitem {ilgun}
Ilgun, K.: USTAT: A Real-time Intrusion Detection System for UNIX. In: Proceedings 1993 IEEE Symposium on Research in Security and Privacy, pp. 16?28. IEEE (1992)
\bibitem {eskin:lee}
Eskin, E., Lee, W., Stolfo, S.J.: Modeling system calls for intrusion detection with dynamic window sizes. In Proceedings DARPA Information Survivability Conference and Exposition II. DISCEX'01, vol. 1, pp. 165?175. IEEE (2001)
\bibitem {ye:xu}
Ye, N., Xu, M., and Emran, S. M.: Probabilistic networks with undirected links for anomaly detection. In: 2000 IEEE Workshop on Information Assurance and Security, West Point, NY (2000)
\bibitem {michael}
Michael, C. C. and Ghosh, A.: Two state-based approaches to program-based anomaly detection. In: ACM Transactions on Information and System Security, no. 5(2), pp. 203?237. ACM, NY (2002)
\bibitem {garvey}
Garvey, T.D., Lunt, T.F.: Model-based Intrusion Detection. In: Proceedings of the 14th Nation computer security conference, vol. 17. Baltimore, MD (1991) 
\bibitem {theus}
Theus, M. and Schonlau, M.: Intrusion detection based on structural zeroes. In: Statistical Computing and Graphics Newsletter, no. 9(1), pp. 12?17 (1998) 
\bibitem {tan}
Tan, K.: The application of neural networks to unix computer security. In: IEEE International Conference on Neural Networks, vol. 1, pp. 476?481, Perth, Australia (1995) 
\bibitem {ilgun:kemmerer}
Ilgun, K., Kemmerer, R.A., Porras, P.A.: State Transition Analysis: A Rule-Based Intrusion Detection System. In: IEEE Trans. Software Eng, vol. 21, no. 3, (1995) 
\bibitem {eskin}
Eskin, E.: Anomaly detection over noisy data using learned probability distributions. In: 17th International Conf. on Machine Learning, pp. 255?262. Morgan Kaufmann, San Francisco, CA (2000)
\bibitem {ghosh}
Ghosh, K., Schwartzbard, A., and Schatz, M.: Learning program behavior profiles for intrusion detection. In: 1st USENIX Workshop on Intrusion Detection and Network Monitoring, pp. 51?62, Santa Clara, California (1999) 
\bibitem {ye}
Ye, N.: A markov chain model of temporal behavior for anomaly detection. In: 2000 IEEE Systems, Man, and Cybernetics, Information Assurance and Security Workshop. IEEE (2000)
\bibitem {axelsson}
Axelsson, S.: The base-rate fallacy and its implications for the difficulty of intrusion detection. In: Proceedings of the 6th ACM conference on Computer and communications security, pp. 1?7. ACM, New York, NY (1999). doi:10.1145/319709.319710 
\bibitem {chikalov}
Chikalov, I, Moshkov, M, Zielosko B.: Optimization of decision rules based on methods of dynamic programming. Vestnik of Lobachevsky State University of Nizhni Novgorod, no. 6, pp. 195?200
\bibitem {chen}
Chen, C.H.: Handbook of pattern recognition and computer vision. University of Massachusetts Dartmouth, USA (2015)
\bibitem {gantmacher}
Gantmacher, F. R.: The theory of matrices. Science, Moscow (1968)
\bibitem {murty}
Murty, M.N., Devi, V.S: Pattern Recognition: An Algorithmic approach, pp. 93?94. Springer Science \& Business Media (2011)
\bibitem {lowe}
Lowe, D.G.: Distinctive Image Features from Scale-Invariant Keypoints, 2004. \url{http://citeseerx.ist.psu.edu/viewdoc/download?doi=10.1.1.157.3843&rep=rep1&type=pdf}
\bibitem {kmeans}
Clustering With K-Means in Python, 2013. \url{https://datasciencelab.wordpress.com/2013/12/12/clustering-with-k-means-in-python}
\bibitem {wu}
Wu, H., Luk, R., Wong, K., and Kwok, K.: Interpreting TF-IDF term weights as making relevance decisions. ACM Transactions on Information Systems, vol. 26, no. 3, (2008)


\end{thebibliography}
\end{document}